\documentclass[10pt,twocolumn,letterpaper]{article}

\usepackage{wacv}              

\usepackage[accsupp]{axessibility}

\usepackage{graphicx}
\usepackage{amsmath}
\usepackage{amssymb}
\usepackage{booktabs}

\usepackage[ruled,vlined]{algorithm2e}
\usepackage{multirow}
\usepackage[normalem]{ulem}
\useunder{\uline}{\ul}{}

\usepackage[pagebackref,breaklinks,colorlinks]{hyperref}

\usepackage[capitalize]{cleveref}
\crefname{section}{Sec.}{Secs.}
\Crefname{section}{Section}{Sections}
\Crefname{table}{Table}{Tables}
\crefname{table}{Tab.}{Tabs.}

\def\wacvPaperID{770} 
\def\confName{WACV}
\def\confYear{2024}

\begin{document}

\title{Distortion-Disentangled Contrastive Learning}

\author{Jinfeng Wang$^{a,b\footnotemark[1]}$, Sifan Song$^{a,b\footnotemark[1]}$, Jionglong Su$^{a\footnotemark[2]}$, S. Kevin Zhou$^{b, c\footnotemark[2]}$\\
$^a$ \small School of AIAC, Xi’an Jiaotong-liverpool University, Suzhou, China\\
$^b$ \small School of BME \& Suzhou Institute for Advanced Research,\\
     \small Center for Medical Imaging, Robotics, Analytic Computing \& Learning (MIRACLE),\\
     \small University of Science and Technology of China, Suzhou, China\\
$^c$ \small Key Lab of Intelligent Information Processing of Chinese Academy of Sciences (CAS),\\
     \small Institute of Computing Technology, CAS, Beijing, China\\
{\tt\small Jionglong.Su@xjtlu.edu.cn}\\
{\tt\small skevinzhou@ustc.edu.cn}
}

\maketitle
\footnotetext[1]{These authors contributed equally to this work.}
\footnotetext[2]{Corresponding authors}
\maketitle

\begin{abstract}
   Self-supervised learning is well known for its remarkable performance in representation learning and various downstream computer vision tasks. Recently, Positive-pair-Only Contrastive Learning (POCL) has achieved reliable performance without the need to construct positive-negative training sets. It reduces memory requirements by lessening the dependency on the batch size. The POCL method typically uses a single objective function to extract the distortion invariant representation (DIR) which describes the proximity of positive-pair representations affected by different distortions. This objective function implicitly enables the model to filter out or ignore the distortion variant representation (DVR) affected by different distortions. However, some recent studies have shown that proper use of DVR in contrastive can optimize the performance of models in some downstream domain-specific tasks. In addition, these POCL methods have been observed to be sensitive to augmentation strategies. To address these limitations, we propose a novel POCL framework named Distortion-Disentangled Contrastive Learning (DDCL) and a Distortion-Disentangled Loss (DDL). Our approach is the first to explicitly and adaptively disentangle and exploit the DVR inside the model and feature stream to improve the representation utilization efficiency, robustness and representation ability. Experiments demonstrate our framework's superiority to Barlow Twins and Simsiam in terms of convergence, representation quality (including transferability and generalization), and robustness on several datasets.
\end{abstract}

\section{Introduction}
\label{sec:intro}

\begin{figure}[h]
\begin{center}
   \includegraphics[width=0.95\linewidth]{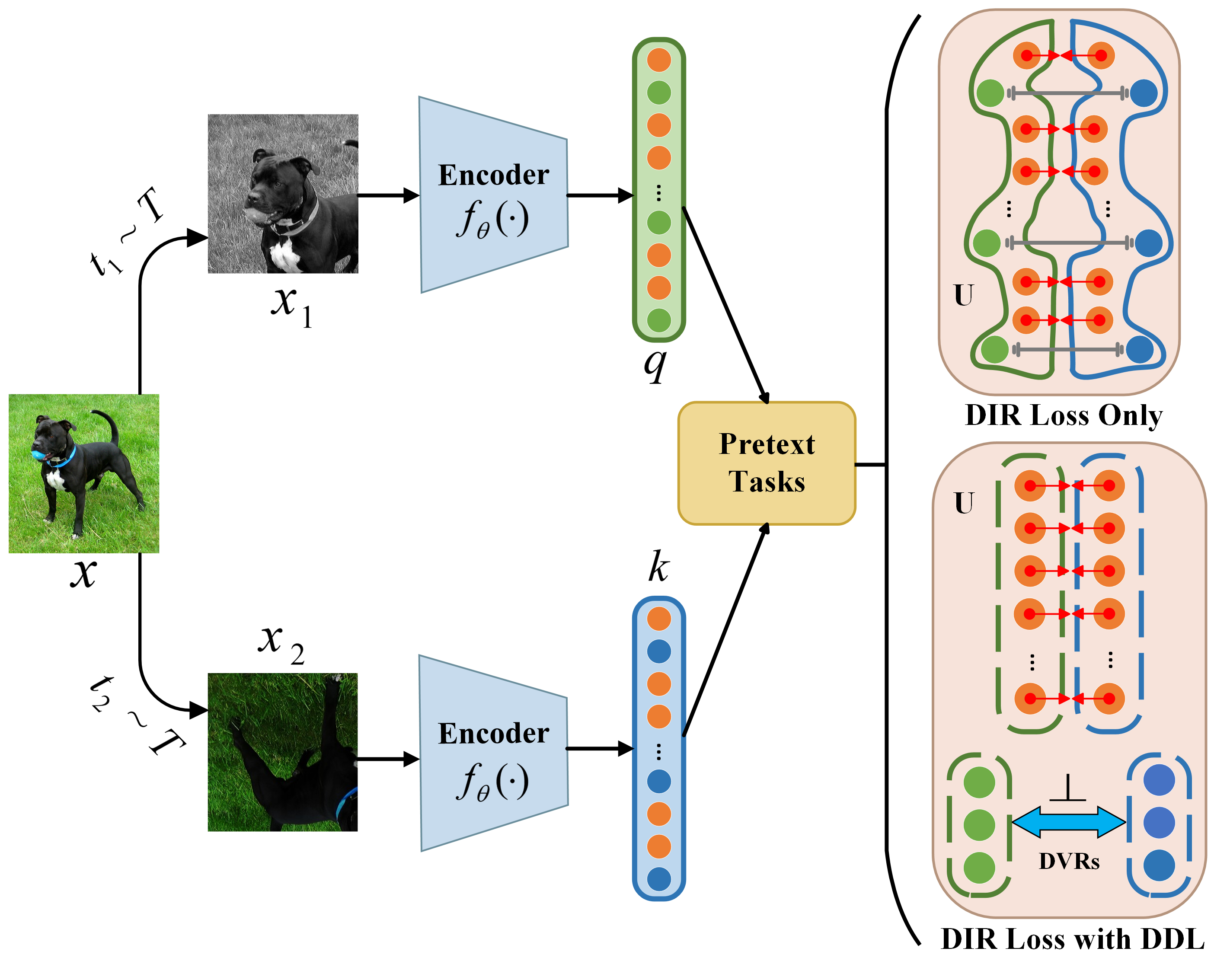}
\end{center}
   \caption{\textbf{U} denotes a two-dimensional feature space constructed using \textit{query} (\textit{q}) and \textit{key} (\textit{k}) based on pretext tasks. The current POCL methods 
   attempt to minimize the distance between the DIRs of positive sample pairs (represented by orange dots) in \textbf{U}. The proposed \textbf{DDL} further ensures that the DVRs for each positive-pair (represented by green and blue dots) are orthogonal within the selected dimension of the overall representation, making them uncorrelated to each other.}
\label{fig:1}
\end{figure}

High-quality representation learning has been a core topic in deep learning research, which is challenging for computer vision due to the low information density \cite{he2020momentum,he2022masked}. In recent years, label-free un/self-supervised contrastive learning methods with instance discrimination as a pretext task undergo steady development, rapidly closing the performance gap with supervised learning methods and demonstrating reliable generalization capabilities \cite{chen2020simple,he2020momentum,grill2020bootstrap,zbontar2021barlow,chen2021exploring}. 

Some previous studies have used different augmented views of the same instance as positive samples with other instances in the mini-batch as negative samples. They employ a series of \textit{query-key} operations to classify samples in the mini-batch and train models to extract the discriminative representations. However, such methods require a large set of positive and negative samples to provide sufficient negative information, such as using large batch sizes \cite{chen2020simple,chen2020big,caron2020unsupervised}, memory banks \cite{wu2018unsupervised,dwibedi2021little}, and memory queues \cite{he2020momentum, chen2020improved,chen2021empirical,tang2022relative}.

Recent POCL methods have achieved superior performance without the use of negative samples \cite{grill2020bootstrap,zbontar2021barlow,chen2021exploring}. These methods aim to minimize the distance between positive sample pairs in the feature space and extract the distortion invariant representation (DIR) by explicitly supervising the Euclidean distance \cite{grill2020bootstrap}, cross-correlation \cite{zbontar2021barlow}, or vector angle \cite{chen2021exploring} of the positive sample pairs. The implicit objective of the POCL methods is to filter out or ignore the distortion variant representation (DVR) using the aforementioned supervision to extract the DIR of the positive sample pairs. However, some recent studies\cite{Xiao_Wang_Efros_Darrell_2021, Devillers_Lefort_2022, dangovski2021equivariant} have shown that the appropriate use of complex distortion strategy and DVR in contrastive learning can improve the performance of models in downstream domain-specific tasks. In addition, the performance of these POCL methods is sensitive to augmentations. As such, augmentation strategies need to be carefully selected for generating positive sample pairs \cite{grill2020bootstrap}. We do observe that these methods have unstable performance on the same test set with different augmentations. Motivated by these, we argue that filtering out or ignoring the DVR using a single loss (\ie, DIR loss) that only supervises the DIR is straightforward but insufficient for representation supervision and utilization in the POCL method. Such inadequately supervised representation may result in the DIR and DVR being entangled in the learned representation, leading to reduced performance.

Therefore, we propose a novel POCL framework, named Distortion-Disentangled Contrastive Learning (DDCL). Unlike previous studies, DDCL does not use augmented information, multi-head structures, or augmentation-specific predictions. This fully adaptive training method makes DDCL applicable to more complex augmentation strategies. In DDCL, we group the last layer (\ie, overall representation) of the encoder into two parts to extract the DIR and DVR. The first part is utilized to extract the DIR using the DIR loss of the corresponding original POCL method. The remaining part is utilized to extract the DVR using our novel loss. We propose a novel Distortion-Disentangled Loss (DDL) to independently supervise the DVR by making the DVRs of a positive sample pair orthogonal. As shown in \cref{fig:1}, DDL explicitly extracts and disentangles the DVR from the overall representation. It is worth noting that the disentangled DVR is not noise or useless information but contains valuable features and distortion information. We further analyze this in \cref{Brick_Study}. By concatenating the DIR and DVR in the following inference task, the model's convergence, representation ability, and robustness are further improved without additional parameters and computation.

We conduct experiments on two existing POCL methods, \ie, Barlow Twins \cite{zbontar2021barlow} and Simsiam \cite{chen2021exploring}. They both use a two-branch architecture and their losses are only designed to extract DIR. Experiments demonstrate that our DDCL and DDL are adaptable to different POCL architectures and corresponding DIR losses, achieving improved performance compared to the original POCL methods. Our main contributions are as follows:

\begin{itemize}
    \item We propose a novel POCL framework, named DDCL, which explicitly disentangles the representation into DIR and DVR. DDCL can adaptively extract the DVR without any other manual designs, which makes DDCL applicable and flexible to more complex augmentation strategies. We can utilize the DVR to enhance the model's performance and robustness.

    \item In DDCL, we propose a novel objective function, DDL, which explicitly supervises and extracts the DVR for its efficient use in the downstream task.

    \item Our proposed DDCL can be adapted to current popular POCL methods. It improves the convergence, representation ability, and robustness without additional inference parameters or computation. 
\end{itemize}


\section{Related Work}

\subsection{Contrastive Learning}
As the core strategy of self-supervised learning, contrastive learning undergoes rapid development in recent years due to its simplicity and efficiency. The main idea of initial contrastive learning \cite{wu2018unsupervised,ye2019unsupervised,he2020momentum,chen2020simple} can be summarized as follows: constructing a set of \textbf{positive and negative samples}; using instance discrimination as a \textbf{pretext task}; and utilizing NCE Loss \cite{gutmann2010noise} or its variants \cite{oord2018representation,yeh2022decoupled} as a \textbf{loss function}. Training an encoder under these settings aims to minimize the distance between positive samples in the feature space while pushing them apart from negative samples. These methods seek sufficient negative samples and appropriate data augmentation strategies for positive samples. Large batch sizes \cite{chen2020simple,chen2020big,caron2020unsupervised}, memory banks \cite{wu2018unsupervised,dwibedi2021little}, memory queues \cite{he2020momentum, chen2020improved,chen2021empirical,tang2022relative}, and clustering structures \cite{caron2020unsupervised,li2022twin,dwibedi2021little,tang2022relative} are utilized to provide sufficient negative sample information. There have been a number of efforts \cite{chuang2020debiased,robinson2020contrastive,shah2022max} to alleviate this negative sample starvation problem in different ways. Recently, some POCL methods that do not use negative samples have been proposed \cite{grill2020bootstrap,zbontar2021barlow,chen2021exploring}. It is worth noting that \cite{Bardes_Ponce_LeCun_2022} cleverly uses other instance information in the mini-batch. These methods do not rely on the constraints of negative samples, and use designed architecture \cite{bromley1993signature}, pretext tasks, loss functions and robustness tricks to prevent model collapse.  

\begin{figure*}[h!t]
\begin{center}
\includegraphics[width = 0.9\linewidth]{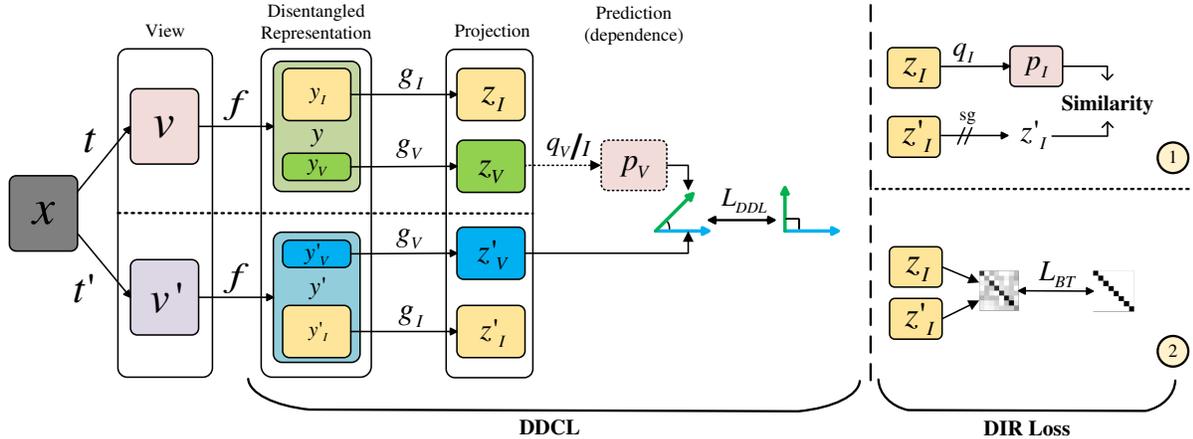}
\end{center}
   \caption{The DDCL framework for symmetric and asymmetric architectures. The encoder denoted as $f$, projector as $g$, and predictor as $q$; the subscripts $I$ and $V$ denote a process or a feature related to DIR and DVR, respectively; $p_V$ and $p_I$ only coexist in the asymmetric version. Two designs of DIR loss correspond to Simsiam \cite{chen2021exploring} and Barlow Twins \cite{zbontar2021barlow}. We explicitly group and disentangle the overall representation into DIR ($y_{I}$ and $y'_{I}$) and DVR ($y_{V}$ and $y'_{V}$) after the encoder, then supervise the mapped features of DIR and DVR using the DIR loss and DDL, respectively.}

\label{fig:2}
\end{figure*}

\subsection{Distortion Variant Contrastive Learning}
Recent studies \cite{Xiao_Wang_Efros_Darrell_2021, dangovski2021equivariant} have shown that the inductive bias of extracting the DIR of an image is not always optimal. The performance of the model on downstream tasks is jointly determined by distortion sensitivity and domain-specific tasks. In order to extract representations that are beneficial for domain-specific tasks in contrastive learning, \cite{Xiao_Wang_Efros_Darrell_2021} designs multiple distortion-dependent prediction heads to obtain multiple distortion-varying subspaces. \cite{dangovski2021equivariant} suggests distortion prediction for a particular operation of distortion to make the model sensitive to that distortion. \cite{Devillers_Lefort_2022} uses the information of the distortion operation to make the model have the specific distortion equivariance in the image space and the feature space. These methods require the manual design of objective functions or model structures for specific distortion operations. This means that these frameworks are pre-trained case-by-case. In contrast, our proposed DDCL can be directly used to adapt a variety of complex distortions without any case-specific modification. Experiments show that DDCL adapts well to affine transformations and elastic transformations.

\subsection{Disentangled Representation}
This topic has become a long-desired goal in the deep learning community subsequently \cite{id1,id2,id3,id4,wang2021self}. The purpose of disentangling is to explicitly decompose the factors of variation from the target representation in a high-dimensional feature space \cite{bengio2013representation,locatello2019challenging}. The disentangled representation has several advantages, including better interpretability \cite{bengio2013representation,li2022interpretable}, utilization efficiency \cite{locatello2019challenging}, robustness \cite{bengio2013representation,wang2018orthogonal,zhang2022decoupled}, and generalization capacity \cite{wang2018orthogonal,cai2019learning,hwang2020variational}. Several works analyze and explicitly use disentangled representations to improve model performance from the perspective of information theory \cite{infor,infor1} and group theory \cite{wang2021self}. Disentangling/Decoupling is also widely used in research fields such as image editing and generation \cite{li2022interpretable}, transfer learning \cite{ge2020zero}, and fairness \cite{locatello2019fairness,sarhan2020fairness}. However, research on disentangled representations in contrastive learning-based self-supervised learning is still in its infancy.

\subsection{Orthogonality}
Orthogonality is usually used in the kernel of deep neural networks to learn more diverse weight matrices and feature vectors \cite{ort1,ort2,li2019orthogonal,wang2020orthogonal}. Several works apply orthogonality in disentangled representation learning \cite{wang2018orthogonal}, model initialization and training \cite{huang2020neural}, and supervised learning with contrastive properties \cite{ranasinghe2021orthogonal}. To the best of our knowledge, our proposed DDCL and DDL are the first to exploit the orthogonality of representations in a contrastive learning task to disentangle distortion information. Unlike previous works on orthogonality, our method does not require additional computations such as singular value decomposition \cite{decom,decom1} and iteration \cite{ite}. The DDL extracts DVR by simply supervising the orthogonality between partial representations.

\section{Proposed Methods}
As mentioned in \cite{dangovski2021equivariant}, the model's sensitivity to certain distortions can effectively improve the feature quality, our purpose of extracting DVR is not to exclude a certain distortion. We hope our model can sensitively capture the heterogeneous features brought by distortions. In contrastive learning with positive and negative samples, many or specific hard negative samples are essential to provide sufficient distortion information and hard samples. In POCL method, only the augmented positive sample pairs provide distortion information for each contrast operation. We find that the performances of existing POCL methods, such as BYOL \cite{grill2020bootstrap}, Barlow Twins \cite{zbontar2021barlow}, and Simsiam \cite{chen2021exploring}, are sensitive to augmentation strategies during training, and become unstable when making inference on unseen distorted inputs. The reason for this may be that POCL methods do not utilize the rare but valuable distortion information. 

\subsection{Revisiting Positive-Only Contrastive Learning}
Existing POCL methods generally comprise of four main factors, \ie, the model architecture, the pretext task, the loss function, and the robustness trick. These POCL methods can be categorized into either symmetric or asymmetric architecture depending on their designs. Barlow Twins (BT) \cite{grill2020bootstrap} employs two entirely symmetric and parameter-sharing encoders. The cross-correlation matrix of the two branches' outputs is used as the pretext task. Furthermore, the distance between the cross-correlation and identity matrix is utilized as the loss function to extract the DIR and eliminate redundancy. Notably, BT uses only batch normalization for robust training. The symmetry loss of BT is as follows:

\begin{equation}
    C = Norm(z)^T \cdot Norm(z')\label{CC}
\end{equation}

\begin{equation}
    Norm(z) = (z-\mu)/\sigma \label{Norm}
\end{equation}

\begin{equation}
    L_{BT} = \sum_i(1-C_{ii})^2 + \lambda \sum_i \sum_{j \neq i} C_{ij}^2 \label{LBT}
\end{equation}
where $C$ refers to the cross-correlation matrix, $z$ and $z'$ refer to the projection of two branches of BT model as shown in \cref{fig:2}. In addition, $C \in \mathbb{R}^{D\times D}$, $z$,$z' \in \mathbb{R}^{B\times D}$, and $\mu,\sigma \in \mathbb{R}^{1\times D}$.

BYOL \cite{grill2020bootstrap} and Simsiam \cite{chen2021exploring} adopt an online-target asymmetric architecture that use regression prediction as the pretext task. BYOL uses the Euclidean distance between the two branches' outputs as the loss function and uses stop-gradient and momentum update strategies to prevent model collapse. Simsiam (Sims) uses the cosine similarity between the two branches' outputs as the loss function and only uses stop-gradient to avoid trivial solutions. From the perspective of model stability and robustness, Simsiam can be regarded as an optimized version based on BYOL. Therefore, we mainly discuss Simsiam in terms of the asymmetric design. The asymmetric loss function is as follows: 

\begin{equation}
    L_{Sims} = \frac{1}{2} {S}(p,sg(z')) + \frac{1}{2} {S}(p',sg(z)) \label{Sims},
\end{equation}
\begin{equation}
    \mathcal{S}(p,z') = - \frac{p\cdot z'}{\parallel p\parallel_2\parallel z'\parallel_2},
\end{equation}
where $p$ refers to prediction, $z$ refers to projection given in \cref{fig:2}-1, $\parallel\cdot\parallel_2$ is an L2-norm, and $sg(\cdot)$ is the stop-grad.

Notably, the primary goal of these POCL methods is to train an encoder that robustly extracts the DIR by narrowing the proximity between different distortion views of the same instance in a high-dimensional feature space, while attempting to ignore or remove the DVR. The loss function of Barlow Twins is designed to explicitly remove redundant information and only retain the DIR, while that of Simsiam is to implicitly ignore the DVR.

We find that, in the original POCL methods that lack distortion information and hard samples, ignoring or eliminating the DVR, either implicitly or explicitly, decreases the overall representation utilization. Moreover, implementing only a single DIR loss to supervise high-dimensional projected representations may be inadequate. Moreover, the models trained using these methods are sensitive to the augmentation strategy during training, making more difficult to infer unseen distorted instances, which leads to reduced robustness. Therefore, we propose the DDCL and DDL to explicitly supervise high-dimensional representations in order to disentangle the DIR and DVR, resulting in the sufficient utilization of concatenated overall representation.

\subsection{Distortion-Disentangled Contrastive Learning}
This paper proposes a novel POCL framework, named Distortion-Disentangled Contrastive Learning (DDCL). When training the model, we group the output of the last layer of each encoder. In this case, the overall representation is grouped into two parts for the DIR and DVR. In addition, the following mapping processes for these two parts are synchronized and supervised by the DIR loss and DDL, respectively. Formulated instructions are as follows: 

\begin{equation}
    y_I = M_{n,I}\cdot f_{1:n-1}(v) \label{DIR}
\end{equation}

\begin{equation}
    y_V= M_{n,V}\cdot f_{1:n-1}(v) \label{DVR}
\end{equation}

\begin{equation}
    y = cat(y_I, y_V) \label{cat} 
\end{equation}

$$M_{n,I} \in \mathbb{R}^{DR\cdot d \times H\cdot W}, M_{n,V} \in \mathbb{R}^{(1-DR)\cdot d \times H\cdot W}$$
where $f_{1:n-1}(\cdot)$ refers to the function of the encoder except the last layer, and $M_n$ refers to the matrix of last layer of the encoder. $y$ refers to the overall representation, and $cat(\cdot)$ is concatenation. $d$ is the output dimension of $f_{1:n-1}$, and $DR$ is the disentangling ratio used to group the overall representation into DIR and DVR parts for separate supervision. 

Since the overall representation ($y$) has been disentangled, the DIR can be used independently for downstream inference tasks. We find that the performance when using only DIR (\ie, $y_I$) is similar to that of the corresponding original POCL method with the full representation. In addition, since the overall representation can be considered as a concatenation of the DIR and DVR, this overall representation achieves even better performance in the subsequent linear evaluation. Furthermore, the overall representation of DDCL is more robust to unseen distorted data. Details of performance are given in \cref{Experiments}. 

\subsection{Distortion-Disentangled Loss}

To supervise the DVR, we propose a novel loss function, named Distortion-Disentangled Loss (DDL). As shown in \cref{fig:2}, the purpose of DDL is to supervise the orthogonality of projected representation vectors, which extracts the DVR from the same instance under different augmentation views. We use the DDL for both symmetric and asymmetric architectures. The formula of DDL in \textbf{symmetric} architecture is as follows: 

\begin{equation}
    \mathcal{D}(z_V, z'_V) \triangleq \mid \frac{z_V\cdot z'_V}{\parallel z_V \parallel_2 \parallel z'_V \parallel_2} - \xi \mid \label{D_sym}
\end{equation}

\begin{equation}
    L_{DDL}^{Sym} \triangleq \mathcal{D}(z_V, z'_V)
\end{equation}
where the hyperparameter $\xi$ is set to 0 in our default setting. The value $z_V$ refers to the projection of DVR (\ie, $y_V$) in \textbf{symmetric} architecture as given in \cref{fig:2}. Therefore, the overall loss in \textbf{symmetric} architecture can be written as:

\begin{equation}
    L^{Sym} = \gamma L_{DDL}^{Sym} + L_{BT}^I
\end{equation}
where $L_{BT}^I$ is the DIR loss of $L_{BT}$ given in \cref{LBT}. The hyperparameter $\gamma$ is set to 1 in our setting. $L_{BT}^I$ is defined as: 
\begin{equation}
    L_{BT}^I \triangleq \sum_i{(1-C_{ii}^I)}^2 + \lambda \sum_i \sum_{j \neq i} {C_{ij}^I}^2 \label{LBTI}
\end{equation}
where
\begin{equation}
    C^I \triangleq Norm(z_I)^T \cdot Norm(z'_I)
\end{equation}

Furthermore, the formulas of DDL in \textbf{asymmetric} architecture is given as follows: 
\begin{equation}
    L_{DDL}^{Asy} \triangleq \frac{1}{2}\mathcal{D}(p_V, sg(z'_V)) + \frac{1}{2}\mathcal{D}(p'_V, sg(z_V))
\end{equation}

\begin{equation}
    L^{Asy} = \gamma L_{DDL}^{Asym} + L_{Sims}^I
\end{equation}

Referring to the definition in \cref{LBTI}, $L_{Sims}^I$ is defined as follows: 

\begin{equation}
    L_{Sims}^I \triangleq \frac{1}{2} {S}(p_I,sg(z'_I)) + \frac{1}{2} {S}(p'_I,sg(z_I))
\end{equation}

As shown by these equations, DDL can be applied to both symmetric and asymmetric architectures. Our design simply groups the overall representation of the last layer of each encoder into two parts. The original loss of the corresponding POCL method is utilized to supervise the DIR part, and DDL is used to supervise the DVR part. Therefore, DDL can be considered to be a plug-in without adding extra dimensions to the overall representation.

\section{Experiments} \label{Experiments}
Based on the Simsiam \cite{chen2021exploring} and Barlow Twins \cite{zbontar2021barlow} methods, we conduct experiments on the convergence, representation quality, and robustness using different scale datasets. These two methods represent the design of asymmetric and symmetric architecture, respectively. In addition, we encounter unstable performance when reproducing BYOL \cite{grill2020bootstrap}. Since Simsiam is more robust and designed for general purposes based on BYOL, our experiments only report the performance of Simsiam. Notably, VICReg \cite{Bardes_Ponce_LeCun_2022} uses other instances in the mini-batch to do contrastive. We cannot ensure it is a POCL method, so we do not explore its performance in this experiment. All reported results in this paper are from our reproductions. Since the overall performance of Simsiam is better than that of Barlow Twins, we only use Simsiam as the baseline in downstream tasks, robustness experiments, and ablation studies. Some related previous work are based on the CL with positive-negative pairs \cite{Xiao_Wang_Efros_Darrell_2021}, and some of them have not been verified by downstream tasks \cite{Devillers_Lefort_2022,dangovski2021equivariant}. No other studies utilize more complex distortions, such as elastic transformation. We believe that the fairest comparison is to compare with baseline methods under various identical distortion settings.

\subsection{Implementation Details}
\textbf{Image augmentations.} This paper reports three augmentation strategies: basic augmentation (BAug), complex augmentation (CAug) and CAug with elastic transformation (CAug$^+$). For the BAug strategy, we refer to the parameters and transformations in Simsiam \cite{chen2021exploring}: random resized cropping, horizontal flipping, color jittering, converting to grayscale, and Gaussian blurring. Similar to \cite{chen2021exploring}, Gaussian blurring is only used to augment the ImageNet datasets (IN-100 and IN-1k) \cite{Deng_Dong_Socher}. In CAug, we add random rotations from -90 to 90 degrees to test the impact of a more complex augmentation strategy on the DDCL and original POCL methods.

\textbf{Architecture.} In small and medium-scale datasets such as CIFAR-10, CIFAR-100 \cite{krizhevsky2009learning}, and STL-10 \cite{coates2011analysis}, we use the Lightly version~\cite{susmelj2020lightly} of ResNet-18 as the backbone. The output dimension of the backbone is 512 in CIFAR experiments and 4608 in STL-10 experiments. In experiments on IN-100 and IN-1k \cite{Deng_Dong_Socher}, we use ResNet-50 as the backbone, and the output dimension of the backbone is 2048. All output dimensions of the following mapping processes, including projection and prediction, are 2048. For fair comparison, the original POCL methods and their corresponding DDCL use the same hyperparameters.

\begin{table}[t] 
\begin{center}
\resizebox{.47\textwidth}{!}
{
\begin{tabular}{lcccccc} 
\hline
\multirow{2}{*}{Method} & \multicolumn{2}{c|}{\textbf{CIFAR-10}}               & \multicolumn{2}{c|}{\textbf{CIFAR-100}}                          & \multicolumn{2}{c}{\textbf{STL-10}} \\ \cline{2-7} 
                        & Top 1          & \multicolumn{1}{c|}{Top 3}          & Top 1                & \multicolumn{1}{c|}{Top 3}                & Top 1            & Top 3            \\ \hline
\multicolumn{7}{c}{100 Epochs}                                                                                                                                                          \\ \hline
Barlow Twins             & {\ul 82.89}    & \multicolumn{1}{c|}{ 96.49}    & 54.98                & \multicolumn{1}{c|}{74.37}                & {\ul 86.90}      & {\ul 97.31}      \\
DDCL\_Sym\_DIR          & 82.78          & \multicolumn{1}{c|}{\ul 96.72}          & {\ul 55.06}          & \multicolumn{1}{c|}{{\ul 74.58}}          & 86.78            & 97.21            \\
DDCL\_Sym               & \textbf{83.64} & \multicolumn{1}{c|}{\textbf{96.81}} & \textbf{56.03}       & \multicolumn{1}{c|}{\textbf{75.31}}       & \textbf{87.10}   & \textbf{97.39}   \\ \hline
SimSiam                 & 75.34          & \multicolumn{1}{c|}{93.68}          & 36.87                & \multicolumn{1}{c|}{56.90}                & {\ul 86.38}      & {\ul 97.46}      \\
DDCL\_Asy\_DIR          & {\ul 76.25}    & \multicolumn{1}{c|}{{\ul 94.36}}    & {\ul 39.77}          & \multicolumn{1}{c|}{{\ul 60.12}}          & 86.32            & 97.44            \\
DDCL\_Asy               & \textbf{76.85} & \multicolumn{1}{c|}{\textbf{94.58}} & \textbf{40.52}       & \multicolumn{1}{c|}{\textbf{61.01}}       & \textbf{86.44}   & \textbf{97.49}   \\ \hline
\multicolumn{7}{c}{200 Epochs}                                                                                                                                                          \\ \hline
Barlow Twins             & 86.17          & \multicolumn{1}{c|}{\textbf{97.46}}          & {\ul 59.60} & \multicolumn{1}{c|}{{\ul 78.34}} & 88.53            & 97.56            \\
DDCL\_Sym\_DIR          & {\ul 86.33}    & \multicolumn{1}{c|}{{97.17}}    & 58.43                & \multicolumn{1}{c|}{77.71}                & {\ul 88.63}      & {\ul 97.66}      \\
DDCL\_Sym               & \textbf{86.72} & \multicolumn{1}{c|}{{\ul 97.44}} & {\textbf{59.61}} & \multicolumn{1}{c|}{{\textbf{78.41}}} & \textbf{89.02}   & \textbf{97.76}   \\ \hline
SimSiam                 & 86.20          & \multicolumn{1}{c|}{97.65}          & {\ul 56.35}          & \multicolumn{1}{c|}{{\ul 77.13}}          & {\ul 89.80}      & {98.17}      \\
DDCL\_Asy\_DIR          & {\ul 87.82}    & \multicolumn{1}{c|}{{\ul 97.91}}    & 56.23                & \multicolumn{1}{c|}{77.05}                & 89.34            & {\ul 98.19}            \\
DDCL\_Asy               & \textbf{88.18} & \multicolumn{1}{c|}{\textbf{98.05}} & \textbf{57.04}       & \multicolumn{1}{c|}{\textbf{78.07}}       & \textbf{89.84}   & \textbf{98.27}   \\ \hline
\end{tabular}
}
\end{center}
\caption{Linear evaluation results of two POCL and the corresponding DDCL architectures pre-trained with 100 and 200 epochs. The best and the second best performance are in bold and underlined, respectively. DIR with a smaller dimension achieves comparable performance compared to the vanilla BT and Sims. Using the overall representation for linear evaluation according to \cref{cat} achieves the best convergence performance.}
\label{Tabel: Con1}
\end{table}

\textbf{Optimization.} Referring to Simsiam \cite{chen2021exploring}, when pre-training the model, we use SGD with base lr = 0.03 on CIFAR and STL-10 with batch size (bs) = 512, base lr = 0.05 on IN-100 (bs = 512) and IN-1k (bs = 256), weight decay = 0.0001, momentum = 0.9, and a cosine decay schedule. For linear evaluation on CIFAR and STL-10, we use an SGD optimizer with 100 epochs, lr = 30.0, weight decay = 0, momentum = 0.9, and batch size = 256. For linear evaluation on IN-100, we use an SGD optimizer with 200 epochs, base lr = 30.0, and batch size = 256. For linear evaluation on IN-1k, we employ a LARS optimizer~\cite{you2017large} with 90 epochs, base lr = 0.1, and batch size = 4096 (similar to Simsiam \cite{chen2021exploring}). 


\subsection{Convergence Study}
We evaluate the performance of the DDCL on CIFAR and STL datasets with small-epoch pre-training (100, 200 epochs). As shown in \cref{Tabel: Con1}, the linear evaluation using only DIR ($y_I$ in \cref{DIR}) achieves approximately on-par convergence performance in smaller dimensions compared to vanilla Barlow Twins and Simsiam. The linear evaluation after concatenating DIR and DVR ($y_V$ in \cref{DVR}) achieves the best convergence performance (\ie, DDCL\_Sym and DDCL\_Asy in \cref{Tabel: Con1}). This suggests that disentanglement and efficient use of the DVR improve the convergence performance of the model.

\begin{table}[t]
\begin{center}
\resizebox{.41\textwidth}{!}
{

\begin{tabular}{llclclcl}
\hline
\multicolumn{2}{l}{Method}                 & \multicolumn{2}{c}{\textbf{CIFAR-10}} & \multicolumn{2}{c}{\textbf{CIFAR-100}} & \multicolumn{2}{c}{\textbf{STL-10}} \\ \hline
\multirow{2}{*}{Barlow Twins}    & Acc. & \multicolumn{2}{c}{87.82}             & \multicolumn{2}{c}{59.66}              & \multicolumn{2}{c}{{\ul 90.68}}     \\
                                & KNN & \multicolumn{2}{c}{84.78}             & \multicolumn{2}{c}{51.63}              & \multicolumn{2}{c}{\textbf{83.61 }}  \\ \hline
\multirow{2}{*}{DDCL\_Sym\_DIR} & Acc. & \multicolumn{2}{c}{{\ul 88.56}}       & \multicolumn{2}{c}{{\ul 59.83}}        & \multicolumn{2}{c}{90.65}           \\
                                & KNN & \multicolumn{2}{c}{/}                 & \multicolumn{2}{c}{/}                  & \multicolumn{2}{c}{/}               \\ \hline
\multirow{2}{*}{DDCL\_Sym}      & Acc. & \multicolumn{2}{c}{\textbf{88.70}}    & \multicolumn{2}{c}{\textbf{60.95}}     & \multicolumn{2}{c}{\textbf{90.83}}  \\
                                & KNN & \multicolumn{2}{c}{\textbf{84.91}}    & \multicolumn{2}{c}{\textbf{51.97}}     & \multicolumn{2}{c}{82.77}           \\ \hline
\multicolumn{8}{l}{}                                                                                                                                              \\ \hline
\multirow{2}{*}{Simsiam}        & Acc. & \multicolumn{2}{c}{91.56}             & \multicolumn{2}{c}{{\ul 66.29}}        & \multicolumn{2}{c}{91.02}           \\
                                & KNN & \multicolumn{2}{c}{87.46}             & \multicolumn{2}{c}{52.38}              & \multicolumn{2}{c}{83.82}           \\ \hline
\multirow{2}{*}{DDCL\_Asy\_DIR} & Acc. & \multicolumn{2}{c}{{\ul 92.01}}       & \multicolumn{2}{c}{65.66}              & \multicolumn{2}{c}{{\ul 91.28}}     \\
                                & KNN & \multicolumn{2}{c}{/}                 & \multicolumn{2}{c}{/}                  & \multicolumn{2}{c}{/}               \\ \hline
\multirow{2}{*}{DDCL\_Asy}      & Acc. & \multicolumn{2}{c}{\textbf{92.19}}    & \multicolumn{2}{c}{\textbf{66.49}}     & \multicolumn{2}{c}{\textbf{91.39}}  \\
                                & KNN & \multicolumn{2}{c}{\textbf{87.90}}    & \multicolumn{2}{c}{\textbf{52.41}}     & \multicolumn{2}{c}{\textbf{84.05}}  \\ \hline
\end{tabular}

}
\end{center}
\caption{Linear evaluation and KNN results of two POCL and the corresponding DDCL architectures pre-trained with 800 epochs. The best and the second best performance are in bold and underlined, respectively. Since the dimensionality of the DIR part is lower than the representation of vanilla methods, we do not report and compare the KNN performance of DDCL\_Sym/Asy\_DIR.}
\label{Tabel: Quality}
\end{table}

\begin{table}[t]
\begin{center}
\resizebox{0.45\textwidth}{!}
{
\begin{tabular}{l|l|l|c|c|c}
\hline
\multicolumn{1}{c|}{Dataset} & \multicolumn{1}{c|}{Train/Epoch} & \multicolumn{1}{c|}{Method} & \multicolumn{1}{c|}{CUB-200}             & \multicolumn{1}{c|}{Flowers-102}        & \multicolumn{1}{c}{Food-101}     \\ \hline
\multirow{9}{*}{IN-100} & \multirow{3}{*}{BAug/500} & Simsiam   & 30.53   & 76.73  & 62.78  \\
& & DIR\_only  & 30.29  & 76.13  & 61.91   \\
& & DDCL\_Asy   & 30.53  & 77.51  & 63.02  \\ \cline{2-6} 
& \multirow{3}{*}{CAug/500} & Simsiam   & 34.79    & 78.57  & 65.37 \\
& & DIR\_only & 34.88    & 78.18  & 64.79    \\
& & DDCL\_Asy    & {\ul 35.11}  & {\ul 78.92}  & {\ul 65.67} \\ \cline{2-6} 
& \multirow{3}{*}{CAug$^+$/500} & Simsiam   & 34.05   & 77.79  & 64.63  \\
& & DIR\_only  & 34.93  & 78.00  & 64.35   \\
& & DDCL\_Asy     & \textbf{35.50}  & \textbf{79.88}  & \textbf{65.74} \\ \hline
\hline
\multirow{6}{*}{IN-1k}& \multirow{3}{*}{CAug/100} & Simsiam  & 39.35 & 79.57 & 71.41 \\
& & DIR\_only & 40.39 & 80.40 & 70.65   \\
& & DDCL\_Asy  & {\ul 40.73} & {\ul 81.98} & 71.79 \\ \cline{2-6} 
& \multirow{3}{*}{CAug/200} & Simsiam  & 40.44 & 80.35 & {\ul 71.93} \\
& & DIR\_only & 40.30 & 80.22 & 71.87   \\
& & DDCL\_Asy  & \textbf{41.34} & \textbf{82.05} & \textbf{72.91} \\ \hline
\end{tabular}
}
\end{center}
\caption{Downstream task performance on domain-specific datasets. Complex distortions and the adaptive DVR extracting method in DDCL significantly improve the transferability and generalization capability.}
\label{Tabel: t&g}
\end{table}

\subsection{Representation Evaluation}
To assess the representation quality extracted by well-trained DDCL models, we evaluate the \textbf{linear evaluation accuracy and KNN performance} of DDCL (DDCL\_Sym and DDCL\_Asy) and the corresponding vanilla POCL methods (Barlow Twins  and Simsiam) after 800 epochs of pre-training in CIFAR and STL datasets. \cref{Tabel: Quality} shows that the classification performance of DIR parts extracted by DDCL (\ie, DDCL\_Sym\_DIR and DDCL\_Asy\_DIR) is still on-par with the vanilla methods after sufficient training. The overall representation extracted by DDCL achieves almost optimal performance in both classification and KNN.

\textbf{Transferability and generalization capability} are essential properties for evaluating the representation quality of models. We pre-train Simsiam and DDCL\_Asy on IN-100 (500 epochs) and IN-1k (100/200 epochs) and then evaluate the transferability and generalization capability in downstream tasks (linear probe) on domain-specific datasets whose distributions are far from ImageNet, such as CUB-200 \cite{WelinderEtal2010}, Flowers-102 \cite{Nilsback2008AutomatedFC}, and Food-101 \cite{bossard14}. As shown in \cref{Tabel: t&g}, DDCL has demonstrated significant improvements on these downstream datasets by utilizing different training distortion strategies and dataset complexities, compared to the original Simsiam with BAug.

\begin{figure*}[t]
\begin{center}
\includegraphics[width = 0.90\linewidth]{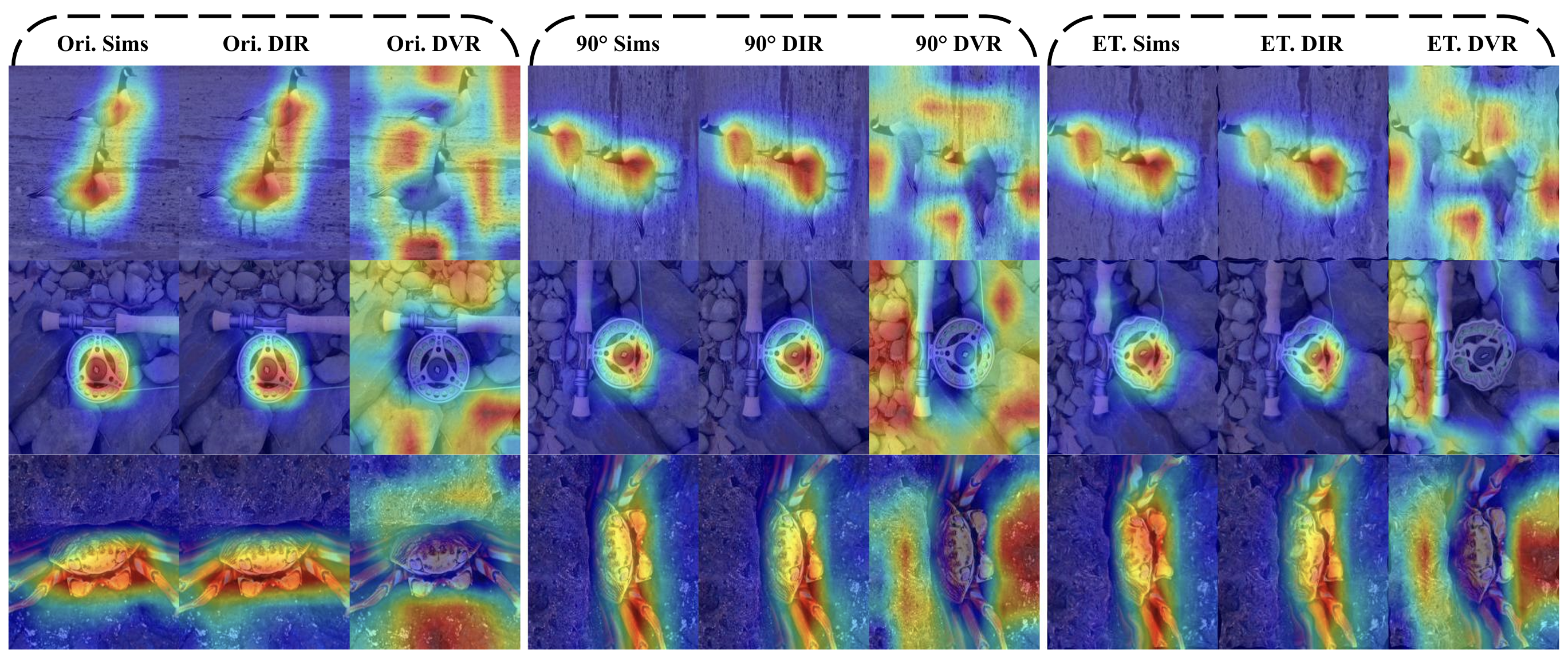}
\end{center}
   \caption{Attention maps of Simsiam and DIR and DVR of DDCL with different distortions on IN-100 (pre-trained using CAug$^+$).}
\label{fig:vis}
\end{figure*}

\begin{table}[t]
\begin{center}
\resizebox{0.47\textwidth}{!}
{

\begin{tabular}{l|l|l|c|c|c|c}
\hline
\multicolumn{1}{c|}{Dataset} & \multicolumn{1}{c|}{Train/Epoch} & \multicolumn{1}{c|}{Method} & \multicolumn{1}{c|}{CJ}             & \multicolumn{1}{c|}{CJ+Flip}        & \multicolumn{1}{c|}{CJ+90°}         & CJ+90°+ET      \\ \hline
\multirow{9}{*}{IN-100} & \multirow{3}{*}{BAug/500} & \multicolumn{1}{l|}{Simsiam}                          & \multicolumn{1}{c|}{{\ul 81.31}}    & \multicolumn{1}{c|}{{\ul 81.40}}    & \multicolumn{1}{c|}{50.18}          & 27.34          \\
&&\multicolumn{1}{l|}{DIR\_only}                   & \multicolumn{1}{c|}{81.24}          & \multicolumn{1}{c|}{81.33}          & \multicolumn{1}{c|}{49.52}          & 26.55          \\
&&\multicolumn{1}{l|}{DDCL\_Asy}                        & \multicolumn{1}{c|}{\textbf{81.60}} & \multicolumn{1}{c|}{\textbf{81.64}} & \multicolumn{1}{c|}{50.02}          & 26.76          \\ \cline{2-7} 
& \multirow{3}{*}{CAug/500} &\multicolumn{1}{l|}{Simsiam}                          & \multicolumn{1}{c|}{78.99}          & \multicolumn{1}{c|}{78.95}          & \multicolumn{1}{c|}{76.95}          & 51.88          \\
&&\multicolumn{1}{l|}{DIR\_only}                   & \multicolumn{1}{c|}{{\ul 79.11}}    & \multicolumn{1}{c|}{{\ul 78.97}}    & \multicolumn{1}{c|}{{\ul 77.29}}    & 49.00          \\
&&\multicolumn{1}{l|}{DDCL\_Asy}                        & \multicolumn{1}{c|}{\textbf{79.33}} & \multicolumn{1}{c|}{\textbf{79.40}} & \multicolumn{1}{c|}{\textbf{77.32}} & 48.49          \\ \cline{2-7} 
& \multirow{3}{*}{CAug$^+$/500} &\multicolumn{1}{l|}{Simsiam}                          & \multicolumn{1}{c|}{\ul 77.67}          & \multicolumn{1}{c|}{\ul 77.69}          & \multicolumn{1}{c|}{75.11}          & 74.06          \\
&&\multicolumn{1}{l|}{DIR\_only}                   & \multicolumn{1}{c|}{{77.64}}    & \multicolumn{1}{c|}{{77.65}}    & \multicolumn{1}{c|}{\textbf{75.44}} & {\ul 74.09}    \\
&&\multicolumn{1}{l|}{DDCL\_Asy}                        & \multicolumn{1}{c|}{\textbf{78.19}} & \multicolumn{1}{c|}{\textbf{78.20}} & \multicolumn{1}{c|}{{\ul 75.37}}    & \textbf{74.27} \\ \hline
\hline
\multirow{6}{*}{IN-1k} & \multirow{3}{*}{CAug/100} &\multicolumn{1}{l|}{Simsiam}  & {\ul 65.59} & {\ul 65.51} & {\ul 62.42} & 28.57 \\
&&\multicolumn{1}{l|}{DIR\_only}  & {65.30} & {65.34} & {62.36} & 28.74 \\
&&\multicolumn{1}{l|}{DDCL\_Asy}  & \textbf{66.07} & \textbf{66.11} & \textbf{62.48} & 29.47 \\ \cline{2-7} 
& \multirow{3}{*}{CAug/200} &\multicolumn{1}{l|}{Simsiam} & {\ul 67.44} & {\ul 67.57} & 64.17 & 31.16 \\
&&\multicolumn{1}{l|}{DIR\_only} & 66.88 & 66.86 & \textbf{64.35} & 30.34 \\
&&\multicolumn{1}{l|}{DDCL\_Asy} & \textbf{67.69} & \textbf{67.79} & {\ul 64.34} & 30.83 \\ \hline
\end{tabular}

}
\end{center}
\caption{Robustness evaluation of models pre-trained by different data augmentation strategies on ImageNet datasets. CJ is random color jetter, 90° denotes randomly applying -90° to 90° rotation, and ET is random elastic transformations ($\alpha=100$, $\sigma=5$).}
\label{Tabel: Robust1}
\end{table}

\subsection{Robustness} \label{Robustness}
Since unseen distortions during pre-training decrease the linear evaluation performance, a common approach to improve model robustness is to apply complex augmentations. We perform a series of experiments to evaluate the robustness of the proposed DDCL. The pre-trained models of Simsiam and DDCL utilized in \cref{Tabel: Robust1} are exactly the same ones used in \cref{Tabel: t&g}.

As shown in \cref{Tabel: Robust1}, when models are pre-trained using only the commonly used augmentation strategy (BAug), the models perform poorly in dealing with rotation distortions (\ie, unseen distortions). After applying a random -90° to 90° rotation to BAug (\ie, CAug), the performance of original POCL (Simsiam) on rotation distortion improves, whereas that on basic distortions decreases by large margins. This performance reduction is justified by the fact that the training process is more difficult (from BAug to CAug) while the training epochs remain unchanged. In contrast, the proposed DDCL has better compatibility with various augmentation strategies. As given in \cref{Tabel: Robust1}, when CAug and CAug$^+$ strategies are used for pre-training, DDCL improves the robustness evaluation performance on rotation and elastic distortions while alleviating performance reduction in basic distortions.


The attention map in \cref{fig:vis} also visually supports this argument and our design goals. The DIR focuses on the region correlated to the target object. It also achieves similar attention performance using a smaller feature dimension (2048$\times$DR) than Simsiam (2048). The DVR focuses on the region that complements the DIR and can further contribute to linear evaluation. In addition, \cref{fig:vis} demonstrates the robustness of our DDCL, as the attention maps of DIR and DVR disentangled by DDCL are highly consistent with different distortions.

\subsection{Ablation Study}

\textbf{Disentangling Ratio.}
The disentangling ratio (DR) mentioned in \cref{DIR} and \cref{DVR} describes the ratio of DIR and DVR in the overall representation. As shown in \cref{Tabel: DR}, the performance of DIR\_only improves as the ratio of DIR increases. Since DR = 0.8 achieves the general best performance for DDCL, we set it as the default value.

\textbf{Batch Size.}
As shown in \cref{table:BS}, DDCL achieves stability and optimal performance at multiple batch sizes. 

\begin{table}[t]
\begin{center}
\resizebox{.43\textwidth}{!}
{
\LARGE
\begin{tabular}{lcc|cc|cc}
\hline
DR  & \multicolumn{2}{c|}{\textbf{CIFAR-10}} & \multicolumn{2}{c|}{\textbf{CIFAR-100}} & \multicolumn{2}{c}{\textbf{STL-10}} \\ \hline
    & DDCL\_Asy          & DIR\_only         & DDCL\_Asy          & DIR\_only          & DDCL\_Asy        & DIR\_only        \\
0.2 & 91.96              & 91.15             & 66.26              & 62.87              & 91.08            & 90.17            \\
0.4 & 92.11              & {\ul 91.58}       & 66.01              & 63.56              & 91.23            & 90.63            \\
0.6 & {\ul 91.72}        & 91.43             & \textbf{66.60}     & {\ul 65.45}        & \textbf{91.47}   & {\ul 91.22}      \\
0.8 & \textbf{92.19}     & \textbf{92.01}    & {\ul 66.49}        & \textbf{65.66}     & {\ul 91.39}      & \textbf{91.28}   \\ \hline
\end{tabular}

}
\end{center}
\caption{Linear evaluation performance with different disentangling ratios (DR).}
\label{Tabel: DR}

\end{table}

\begin{table}[]
\begin{center}
\resizebox{.39\textwidth}{!}
    {

\begin{tabular}{lccccc}
\hline
Method\textbackslash{}Batch Size & 32    & 64    & 128   & 256   & 512   \\ \hline
Simsiam                          & 91.66 & 91.44 & 91.25 & 90.85 & 91.56 \\
DIR\_only                        & 91.60 & 91.51 & 90.96 & 91.41 & 92.01 \\
DDCL\_Asy                        & \textbf{91.75} & \textbf{91.66} & \textbf{91.26} & \textbf{91.64} & \textbf{92.19} \\ \hline
\end{tabular}
    
    }
\end{center}

\caption{Linear evaluation on CIFAR-10 with different batch sizes.}
\label{table:BS}
\end{table}

\subsection{Brick Study} \label{Brick_Study}
We design this novel brick study to further explore the impact of DVR on the overall representation. As the DIR and DVR parts have been grouped and disentangled from the overall representation, they can be utilized independently or concatenated to others (as bricks). Therefore, in \cref{Tabel: brik}, we concatenate the DIR (row) with the DVR (column) disentangled from various instances and distortions, and evaluate the linear classification performance of this new overall representation. This part of the research is carried out on the STL-10 dataset.

Based on the performance difference observed between two augmentation strategies when facing rotation (columns 2 and 3 in \cref{Tabel: brik}), we argue that DVR can extract content features when encountering the unseen distortion that the model cannot handle (third row). When the model is trained with the rotation distortion (\ie, CAug), DVR can mainly extract the distortion-related features as designed. However, the current DVR cannot clearly decompose the content and distortion-related features.  

\begin{table}[t]
\begin{center}
\resizebox{.45\textwidth}{!}
{
\LARGE

\begin{tabular}{lccccc}
\hline
\multicolumn{1}{l|}{DIR\textbackslash{}DVR} & \multicolumn{1}{c|}{Orig.}  & \multicolumn{1}{c|}{Flip}  & \multicolumn{1}{c|}{Flip+90°} & \multicolumn{1}{c|}{Dif.Inst+Flip+90°} & Zero DVR \\ \hline
\multicolumn{6}{c}{Trained by BAug}                                                                                                                                                         \\ \hline
\multicolumn{1}{l|}{Orig.}                   & \multicolumn{1}{c|}{\textbf{91.39}} & \multicolumn{1}{c|}{91.31} & \multicolumn{1}{c|}{90.76}    & \multicolumn{1}{c|}{90.34}             & 89.78    \\
\multicolumn{1}{l|}{Flip}                   & \multicolumn{1}{c|}{91.44} & \multicolumn{1}{c|}{\textbf{91.32}} & \multicolumn{1}{c|}{90.76}    & \multicolumn{1}{c|}{90.32}             & 89.95    \\
\multicolumn{1}{l|}{Flip+90°}               & \multicolumn{1}{c|}{62.58} & \multicolumn{1}{c|}{62.69} & \multicolumn{1}{c|}{\textbf{49.85}}    & \multicolumn{1}{c|}{46.44}             & 48.46    \\ \hline
\multicolumn{6}{c}{Trained by CAug}                                                                                                                                                         \\ \hline
\multicolumn{1}{l|}{Orig.}                   & \multicolumn{1}{c|}{\textbf{89.98}} & \multicolumn{1}{c|}{89.97} & \multicolumn{1}{c|}{88.89}    & \multicolumn{1}{c|}{88.89}             & 88.41    \\
\multicolumn{1}{l|}{Flip}                   & \multicolumn{1}{c|}{89.69} & \multicolumn{1}{c|}{\textbf{89.69}} & \multicolumn{1}{c|}{88.90}    & \multicolumn{1}{c|}{88.56}             & 88.42    \\
\multicolumn{1}{l|}{Flip+90°}               & \multicolumn{1}{c|}{87.20} & \multicolumn{1}{c|}{87.32} & \multicolumn{1}{c|}{\textbf{84.92}}    & \multicolumn{1}{c|}{84.66}             & 84.58    \\ \hline
\end{tabular}

}
\end{center}
\caption{`Dif.Inst' means that the DVR and DIR come from different instances, and `Zero DVR' means to set the DVR part to zero values. The bold numbers represent the DIR and DVR originating from the same pre-training model (\ie, without any altered bricks).}
\label{Tabel: brik}

\end{table}

In the columns `Flip+90°' and `Dif.Inst+ Flip+90°' (\cref{Tabel: brik}), the linear evaluation performance of concatenated DIR and DVR is generally comparable when the DVR is generated by the same kind of distortion, regardless of whether the DVR and DIR come from the same instance. This is due to the fact that the DVR contains certain information representing the distortion itself.

As shown in the third row of \cref{Tabel: brik}, since the model is trained with the BAug strategy, rotation is an unseen distortion that cannot be handled. The DVR may extract content features, so the zero DVR at this time does not significantly impact the performance (third row, fifth column), which is consistent with our purpose. When using the DVR of other instances to classify this instance (third row, fourth column), the classification performance further deteriorates. This performance reduction may be caused by the influence of content features from other instances.

\section{Conclusion and Discussion}
This paper proposes a novel POCL framework, DDCL, and a novel objective function, DDL, to adaptively extract the DVR part from the overall representation. We apply the DDCL to both symmetric and asymmetric POCL architectures to improve model convergence, representation quality, and robustness by explicitly supervising and adaptively disentangling the DVR inside the model. Meanwhile, we analyze the composition of the DVR through a novel brick study. For DDCL, we plan to extend this design to positive and negative sample contrastive learning frameworks to explore the potential of adaptive DVR extracting. Furthermore, we believe that the information in the DVR is worthy of further analysis and even subsequent disentangling. In addition to this, the role of DVRs in dense prediction tasks such as segmentation and target detection is also very promising. We will further explore these in our future work.

\section*{Acknowledgment}

This work was supported by the Key Program Special Fund in XJTLU (KSF-A-22). Zhou is supported by National Natural Science Foundation of China (NSFC) under grant No. 62271465.

{\small
\bibliographystyle{ieee_fullname}
\bibliography{egbib}
}

\newpage
\clearpage

\renewcommand\thesection{\Alph{section}}
\numberwithin{table}{section}
\numberwithin{figure}{section}
\setcounter{section}{0}
\setcounter{table}{0}
\setcounter{page}{1}
\section*{SUPPLEMENTARY MATERIAL}

\section{Pseudocode}

\definecolor{commentcolor}{RGB}{110,154,155}
\newcommand{\PyComment}[1]{\ttfamily\footnotesize\textcolor{commentcolor}{\# #1}} 
\newcommand{\PyCode}[1]{\ttfamily\footnotesize\textcolor{black}{#1}}

\begin{algorithm}[h]
    \SetAlgoLined
    \caption{DDCL Pseudocode, PyTorch-like}
    \label{algo: DDCL}
    \hspace{0pt}\\
    
    \PyComment{f: encoder} \\
    \PyComment{g: projection mlp} \\
    \PyComment{q: prediction mlp} \\
    \PyComment{\_I, \_V: DIR, DVR} \\
    \PyComment{DR: disentangling ratio}\\
    \PyComment{off\_diagonal: off-diagonal elements}\\
    \hspace{0pt}\\

    \PyCode{for x in loader:} \\
    \Indp
        \PyComment{two randomly augmented versions of x}\\
        \PyCode{x1, x2 = aug(x), aug(x)}\\ 
        \PyCode{y1, y2 = f(x1), f(x2)} \PyComment{representations}\\
        \hspace{0pt}\\
        
        \PyCode{d\_I = y1.size(1)*DR} \PyComment{DIR dimension}\\
        \hspace{0pt}\\

        \PyComment{grouping and disentangling}\\
        \PyCode{y1\_I, y2\_I = y1[:,:d\_I], y2[:,:d\_I]} \PyComment{DIR}\\
        \PyCode{y1\_V, y2\_V = y1[:,d\_I:], y2[:,d\_I:]} \PyComment{DVR}\\
        \hspace{0pt}\\

        \PyComment{projections for DIR and DVR}\\
        \PyCode{z1\_I, z2\_I = g\_I(y1\_I), g\_I(y2\_I)}\PyComment{NxD}\\
        \PyCode{z1\_V, z2\_V = g\_V(y1\_V), g\_V(y2\_V)}\PyComment{NxD}\\
        \hspace{0PT}\\

        \PyComment{------if method is Barlow Twins------}\\

        \PyComment{Batch Normalization}\\
        \PyCode{z1\_I, z2\_I} = BN(Z1\_I), BN(z2\_I)\\
        \PyCode{z1\_V, z2\_V} = BN(Z1\_V), BN(z2\_V)\\
        \hspace{0PT}\\
        
        \PyComment{Cross-Corr matrix}\\
        \PyCode{C = mm(z1\_I.T, z2\_I)/N} \PyComment{DxD}\\  
        \PyCode{C\_diff = (C - eye(C.size(0))).pow(2)} \\
        \PyCode{off\_diagonal(C\_diff).mul\_(lambda)}\\
        \hspace{0pt}\\
        
        \PyCode{L\_BT = C\_diff.sum()} \PyComment{loss of BT}\\
        \PyCode{DDL\_sym = D(z1\_V, z2\_V)}\\
        \PyCode{L\_sym = gamma*DDL\_sym + L\_BT(z1\_I, z2\_I)}\\
        \hspace{0pt}\\

        \PyComment{--------if method is Simsiams--------}\\
        \PyComment{predictions}\\
        \PyCode{p1\_I, p2\_I = q\_I(z1\_I), q\_I(z2\_I)}\\
        \PyCode{p1\_V, p2\_V = q\_V(z1\_V), q\_I(z2\_V)}\\
        \hspace{0pt}\\
        
        \PyComment{Stop-grad}\\
        \PyCode{z1\_I, z2\_I = z1\_I.detach(), z2\_I.detach()}\\
        \PyCode{z1\_V, z2\_V = z1\_V.detach(), z2\_V.detach()}\\
        \hspace{0pt}\\
        
        \PyCode{S1 = -cos\_simiarity(p1\_I, p2\_I)}\\
        \PyCode{S2 = -cos\_simiarity(p2\_I, p1\_I)}\\
        \hspace{0pt}\\
        
        \PyCode{L\_Sims = S1/2 + S2/2} \PyComment{loss of Sims}\\
        \PyCode{DDL\_asy = D(p1\_V,z2\_V)/2+D(p2\_V,z1\_V)/2}\\
        \PyCode{L\_asy = gamma*DDL\_asy + L\_Sims}\\
        \hspace{0pt}\\
    \Indm
    
\PyCode{def D(r1, r2):}\\
    \Indp
        return abs(cos\_similarity(r1, r2) - eps)\\
    \Indm
    \hspace{0pt}\\

\end{algorithm}

\newpage

\section{Robustness in STL-10}
\begin{table}[h]
\begin{center}
\resizebox{.45\textwidth}{!}
{

\begin{tabular}{lcccc}
\hline
\multicolumn{1}{l|}{Method $\backslash$ Distortion}        & \multicolumn{1}{l|}{CJ}             & \multicolumn{1}{l|}{CJ + Flip}      & \multicolumn{1}{l|}{CJ + 90°}       & \multicolumn{1}{l}{CJ + 180°} \\ \hline
\multicolumn{5}{c}{Trained by BAug}                                                                                                                                                     \\ \hline
\multicolumn{1}{l|}{BT}             & \multicolumn{1}{c|}{90.62}          & \multicolumn{1}{c|}{90.56}          & \multicolumn{1}{c|}{\textbf{43.29}} & \textbf{37.10}                \\
\multicolumn{1}{l|}{DDCL\_Sym\_DIR} & \multicolumn{1}{c|}{90.43}          & \multicolumn{1}{c|}{90.47}          & \multicolumn{1}{c|}{41.07}          & 34.87                         \\
\multicolumn{1}{l|}{DDCL\_Sym}      & \multicolumn{1}{c|}{90.74}          & \multicolumn{1}{c|}{90.64}          & \multicolumn{1}{c|}{ 41.70}    & 35.61                         \\
\multicolumn{1}{l|}{Simsiam}        & \multicolumn{1}{c|}{90.84}          & \multicolumn{1}{c|}{90.94}          & \multicolumn{1}{c|}{42.34}          & 36.56                         \\
\multicolumn{1}{l|}{DDCL\_Asy\_DIR} & \multicolumn{1}{c|}{{\ul 91.06}}    & \multicolumn{1}{c|}{{\ul 91.12}}    & \multicolumn{1}{c|}{42.25}          & 36.15                         \\
\multicolumn{1}{l|}{DDCL\_Asy}      & \multicolumn{1}{c|}{\textbf{91.19}} & \multicolumn{1}{c|}{\textbf{91.22}} & \multicolumn{1}{c|}{{\ul 42.36}}          & {\ul 36.72}                   \\ \hline
\multicolumn{5}{c}{Trained by CAug}                                                                                                                                                     \\ \hline
\multicolumn{1}{l|}{BT}             & \multicolumn{1}{c|}{87.76}          & \multicolumn{1}{c|}{87.80}          & \multicolumn{1}{c|}{85.23}          & 80.79                         \\
\multicolumn{1}{l|}{DDCL\_Sym\_DIR} & \multicolumn{1}{c|}{87.57}          & \multicolumn{1}{c|}{87.32}          & \multicolumn{1}{c|}{85.55}          & 80.90                         \\
\multicolumn{1}{l|}{DDCL\_Sym}      & \multicolumn{1}{c|}{88.29}          & \multicolumn{1}{c|}{88.32}          & \multicolumn{1}{c|}{85.75}          & 80.82                         \\
\multicolumn{1}{l|}{Simsiam}        & \multicolumn{1}{c|}{88.52}          & \multicolumn{1}{c|}{88.37}          & \multicolumn{1}{c|}{86.63}          & 82.18                         \\
\multicolumn{1}{l|}{DDCL\_Asy\_DIR} & \multicolumn{1}{c|}{{\ul 89.26}}    & \multicolumn{1}{c|}{{\ul 89.38}}    & \multicolumn{1}{c|}{{\ul 87.47}}    & {\ul 82.28}                   \\
\multicolumn{1}{l|}{DDCL\_Asy}      & \multicolumn{1}{c|}{\textbf{89.57}} & \multicolumn{1}{c|}{\textbf{89.48}} & \multicolumn{1}{c|}{\textbf{87.55}} & \textbf{82.69}                \\ \hline
\end{tabular}

}
\end{center}
\caption{Robustness evaluation by different augmentation strategies on STL-10. 90° and 180° denote randomly applying -90° to 90° and -180° to 180° rotation during inference, respectively.}
\label{Tabel: Robust1}

\end{table}

\section{Ablation Study on Warm-up}

\begin{table}[h]
\begin{center}
\resizebox{.45\textwidth}{!}
{

\begin{tabular}{ccl|cl|cl}
\hline
\multicolumn{1}{l}{Warm-up} & \multicolumn{2}{c|}{\textbf{CIFAR-10}}                          & \multicolumn{2}{c|}{\textbf{CIFAR-100}}                         & \multicolumn{2}{c}{\textbf{STL-10}}                   \\ \hline
\multicolumn{1}{l}{}        & \multicolumn{1}{l}{w/o}   & w/30                                & \multicolumn{1}{l}{w/o}   & w/30                                & \multicolumn{1}{l}{w/o}   & w/30                      \\
Simsiam                     & 91.83                     & \multicolumn{1}{c|}{91.56}          & 65.99                     & \multicolumn{1}{c|}{66.29}          & 91.22                     & \multicolumn{1}{c}{91.02} \\
DIR\_only                   & \multicolumn{1}{l}{91.75} & 92.01                               & \multicolumn{1}{l}{65.77} & 65.66                               & \multicolumn{1}{l}{91.26} & 91.28                     \\
DDCL\_Asy                   & 91.84                     & \multicolumn{1}{c|}{\textbf{92.19}} & 66.20                     & \multicolumn{1}{c|}{\textbf{66.49}} & \textbf{91.43}            & \multicolumn{1}{c}{91.39} \\ \hline
\end{tabular}
}
\end{center}
\caption{Effect of warm-up on the linear evaluation.}
\label{Tabel: WP}

\end{table}

\begin{figure}[h!]
\begin{center}
   \includegraphics[width=\linewidth]{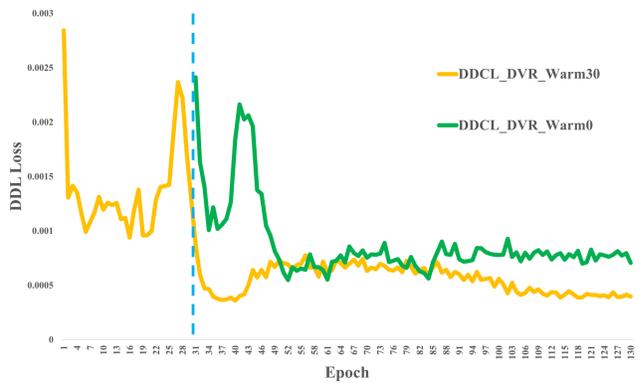}
\end{center}
\caption{The influence of Warm-up on Distortion-Disentangled Loss in DDCL\_Asy when pre-training.}
\label{fig:WP}
\end{figure}

\textbf{Warm-up}  
We use a warm-up of 30 epochs in the default design to accompany the cosine training schedule. As shown in \cref{Tabel: WP} and \cref{fig:WP} , the overall performance of DIR\_only and DDCL\_Asy is generally improved by the warm-up strategy, and outperforms that of Simsiam.

{}

\renewcommand{\thefigure}{D.1}
\begin{figure*}[t]
\begin{center}
\includegraphics[width = \linewidth]{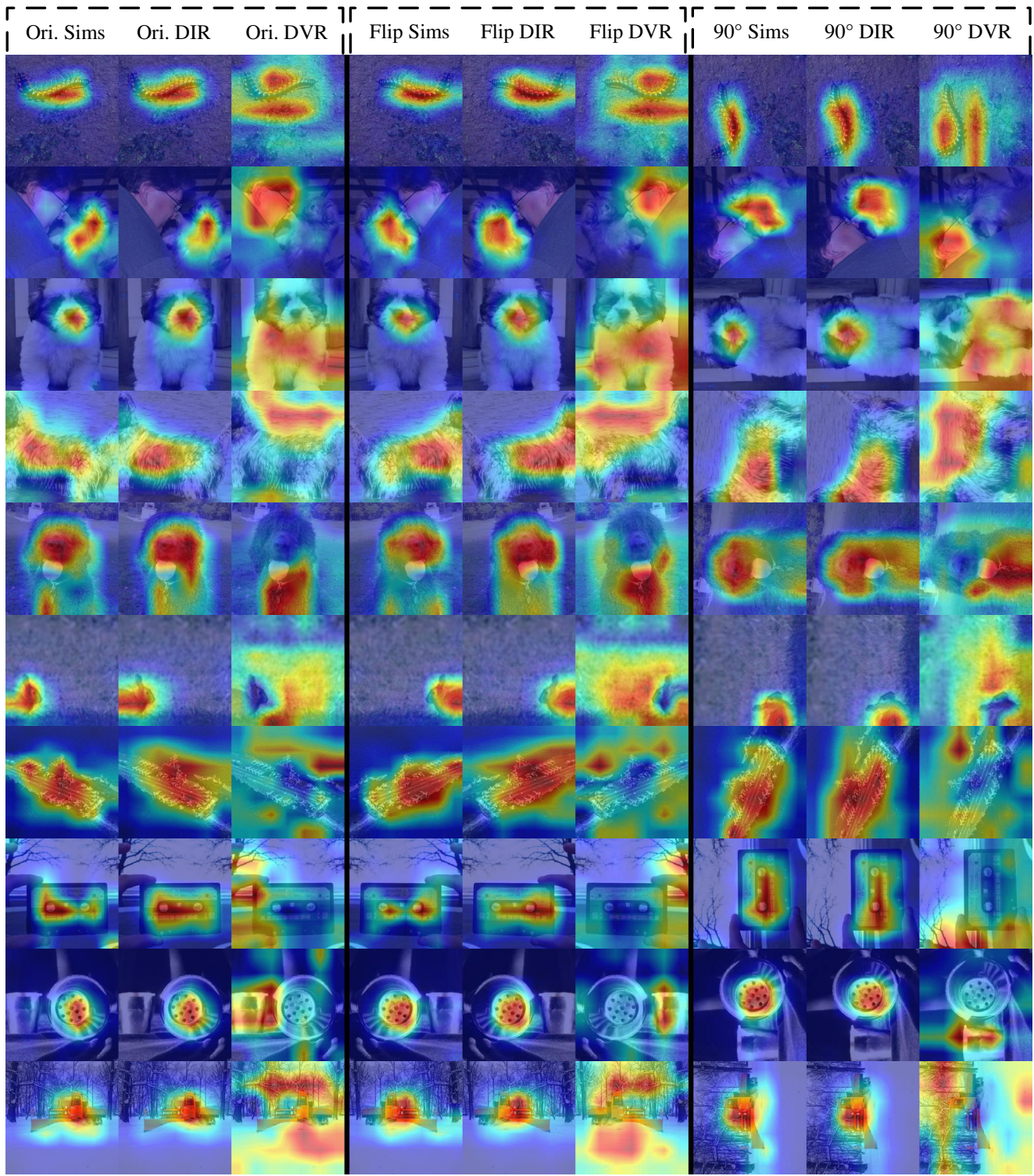}
\end{center}
   \caption{In order to analyze the semantics of DVR, we report more attention maps of Simsiam and DIR and DVR of DDCL Asy of inputs with different distortions on ImageNet (pre-trained using CAug). ‘Ori.’, ‘Flip’ and ‘90°’ represent that the input is original, horizontally flipped, and rotated by 90°. }
\label{fig:A1}
\end{figure*}

\section{Semantic Analysis of DVR}

We analysis more attention maps to study the semantics referred to by DVR as shown in Figure \ref{fig:A1}. The semantics of DVR's regions of interest can be considered: the environment background, some components of the object, other foregrounds, and the "envelope" of the target object. 

It is easy to understand that information about the environmental background and some components of the target object can provide additional information to improve the performance of the prediction (perhaps involving fairness analysis). The information of other foregrounds may give the model potential for multi-object detection and robustness to label noise (needs further exploration). In addition, we argue that this kind of "envelope" wrapping the target object can bring certain distortion information to the model (\ie, the rotation and flipping distortion in the figure).

\end{document}